\documentclass[letterpaper, 10 pt, conference]{ieeeconf}  

\IEEEoverridecommandlockouts                              

\usepackage{graphicx} 
\usepackage{amsmath} 
\usepackage{amssymb}  
\usepackage{mathtools}
\usepackage{bm} 
\usepackage{xcolor}
\usepackage{booktabs}
\usepackage{multirow}

\newcommand{\red}[1]{\textcolor{red}{}} 

\newcommand{\blue}[1]{\textcolor{blue}{}} 

\title{\LARGE \bf
Improving Cross-embodiment Transfer in Latent Action Models with Action-Similarity Supervision
}

\author{Maxime Alvarez$^{1,*}$, Renzo Caballero$^{1}$, Tatsuya Matsushima$^{1}$, Yusuke Iwasawa$^{1}$, Yutaka Matsuo$^{1}$%
\thanks{$^{1}$Graduate School of Engineering,
        The University of Tokyo%
}
\thanks{
        $^{*}$ Corresponding author: {\tt maxime.alvarez@weblab.t.u-tokyo.ac.jp}%
}
}

\begin{document}

\maketitle
\thispagestyle{empty}
\pagestyle{empty}

\begin{figure*}
    \centering
    \includegraphics[width=\linewidth]{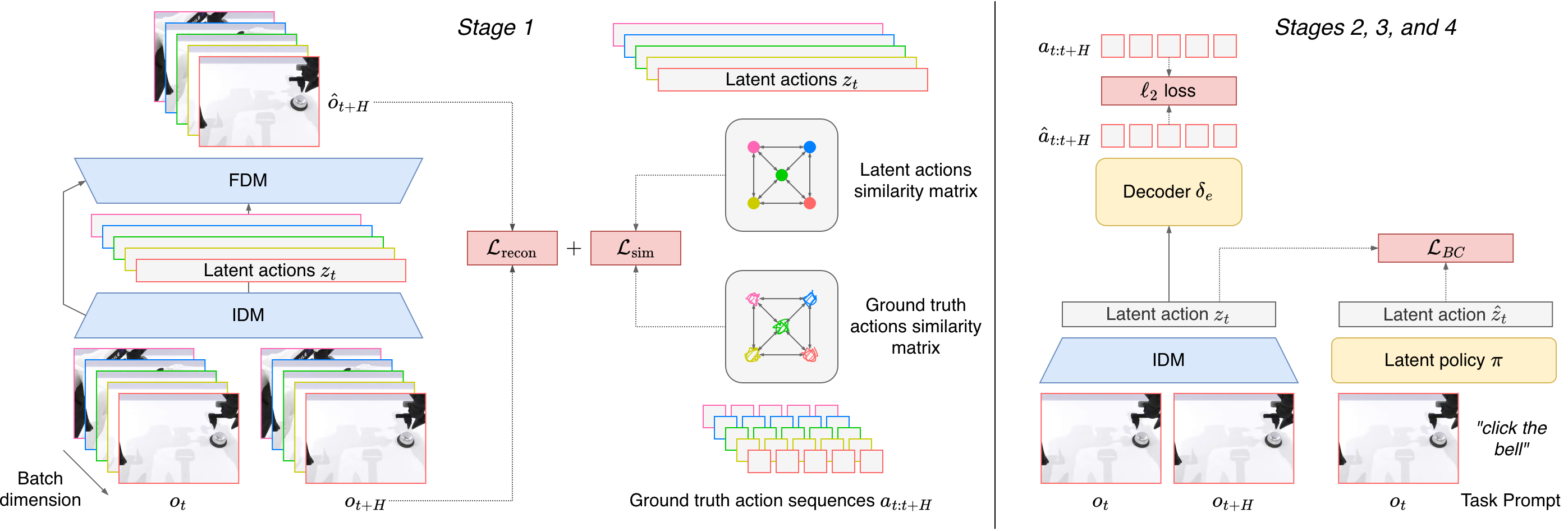}
    \caption{Illustration of our proposed method. During the first stage, we sample a batch of observations $o_t$ and future observations $o_{t+H}$. The inverse dynamics model (IDM) encodes each pair into a single low-dimensional latent action $z_t$. Within a batch, we compute the pairwise similarities of the latent actions and align them with the pairwise similarities of the corresponding ground-truth action sequences ($\mathcal{L}_\text{sim}$). The forward dynamics model (FDM) reconstructs the future observation from the current observation $o_t$ and the latent action ($\mathcal{L}_\text{recon}$). The IDM and FDM training is done jointly by optimizing the combined loss $\mathcal{L} = \mathcal{L}_\text{recon} + \lambda\mathcal{L}_\text{sim}$. In the second stage we annotate the dataset using the trained IDM, then in the third stage we train the latent behavior cloning policy ($\pi$ optimizing $\lambda\mathcal{L}_\text{BC}$), finally in stage 4 we train a decoder $\delta_e$ for each embodiment $e$, using observations pairs associated with their ground-truth actions ($a_{t:t+H}$) and the corresponding latent action. In practice stage 3 and 4 are done in parallel as they are independent from each other. At inference, observations are fed to the latent policy $\pi$ and its output is processed by the appropriate decoder $\delta_e$ before being executed on the robot.}
    \label{fig:main_illustration}
\end{figure*}

\begin{abstract}

As generalist robot policies gain vision and language from web-scale pretraining, demonstrations remain costly to collect and tied to the robot that recorded them. Latent action models (LAMs) address both by learning latent actions from action-free videos that can be shared across embodiments, however, in practice, LAMs are sensitive to background visual noise, and the same motion from two different robots may be encoded with different latents. One solution to the background visual noise is to add an auxiliary loss predicting the robot action from the latent action, further associating the latent action space to the embodiment specific robot action space. We study a different use of the same labels, through action-similarity supervision. The similarity between any two latent actions is trained to match the similarity of the two ground-truth robot action sequences. The ground-truth actions are never predicted by the LAM, so the latent action does not need to encode embodiment specifics. We evaluate cross-embodiment transfer on RoboTwin 2.0 in a controlled setup, two bimanual robots demonstrate disjoint task sets, a policy is trained on all the demonstrations, and each robot is evaluated closed-loop on the tasks only the other demonstrated. With the policy architecture and its hyperparameters, the dataset, and the evaluation protocol fixed, predicting latent actions instead of ground-truth actions more than doubles cross-embodiment success. Given the same ground-truth actions, similarity supervision transfers better than an auxiliary loss that predicts the ground-truth action during the LAM training. Computing the similarities on end-effector motion rather than joint-space motion, and letting the loss compare latent actions across the two robots, gives the best approach of the study.

\end{abstract}

\section{Introduction}
The rapid progress of generalist robotic policies, such as vision-language-action (VLA) models, has been fueled by the use of vision-language models (VLM) trained on internet-scale datasets \cite{black2024pi0visionlanguageactionflowmodel,  intelligence2025pi05visionlanguageactionmodelopenworld, shukor2025smolvlavisionlanguageactionmodelaffordable, nvidia2025gr00tn1openfoundation}. Although the introduction of such models has greatly improved the vision and language capabilities of robotic policies, while some benchmarks have been practically solved, a number of them still remain challenging \cite{yang2026vlact}. One reason is that, while we have scaled the vision-language data, the number of expert robotic trajectories containing actions has been limited due to the cost and difficulties related to their collection \cite{khazatsky2024droid, mandlekar2023mimicgen}. In that context, latent action models (LAM), have been proposed as an alternative to expensive action-labeled expert trajectories by \textit{leveraging} trajectories collected without action labels and then being able to learn directly from large-scale videos datasets from the web \cite{schmidt2023lapo, bruce2024geniegenerativeinteractiveenvironments, chen2024moto, ye2024latentactionpretrainingvideos}. Most LAMs still require a limited amount of action-labeled data to be able to predict real actions, which is done either by fine-tuning a latent policy \cite{ye2024latentactionpretrainingvideos} or by learning a small network that will decode latent actions into real actions \cite{schmidt2023lapo}. LAMs are typically built in multiple stages that are inverse dynamic model (IDM) and forward dynamic model (FDM) joint pre-training using observations only, data annotation using the IDM, latent policy training, and decoding training using a small amount of ground-truth action labeled trajectories.
Using only observation to train the LAM introduces challenges such as differentiating external noise and causal changes \cite{yeom2026actionrelevant}, or learning similar actions across visually different embodiments. Most importantly, by not leveraging action labels during the initial stages, they are missing important clues provided by real actions when available. Some recent works have proposed using the action-labeled transitions with an auxiliary loss directly predicting the true action from the latent action \cite{nikulin2025latent, zhang2026latent}, or using contrastive learning and similarity matching approaches on the latent action space guided by the ground-truth action space \cite{jeong2026vila}.

In this work, we study how to leverage these action-labeled trajectories to improve cross-embodiment transfer in LAMs, and demonstrate that a simple action-similarity supervision during the latent action pretraining transfers up to 88.5\% of the in-embodiment performance to a new embodiment, illustrated in Figure~\ref{fig:main_illustration}, doubling the cross-embodiment success rate on our benchmark compared to $\pi_0$~\cite{black2024pi0visionlanguageactionflowmodel}.

\begin{figure}
    \centering
    \includegraphics[width=\linewidth]{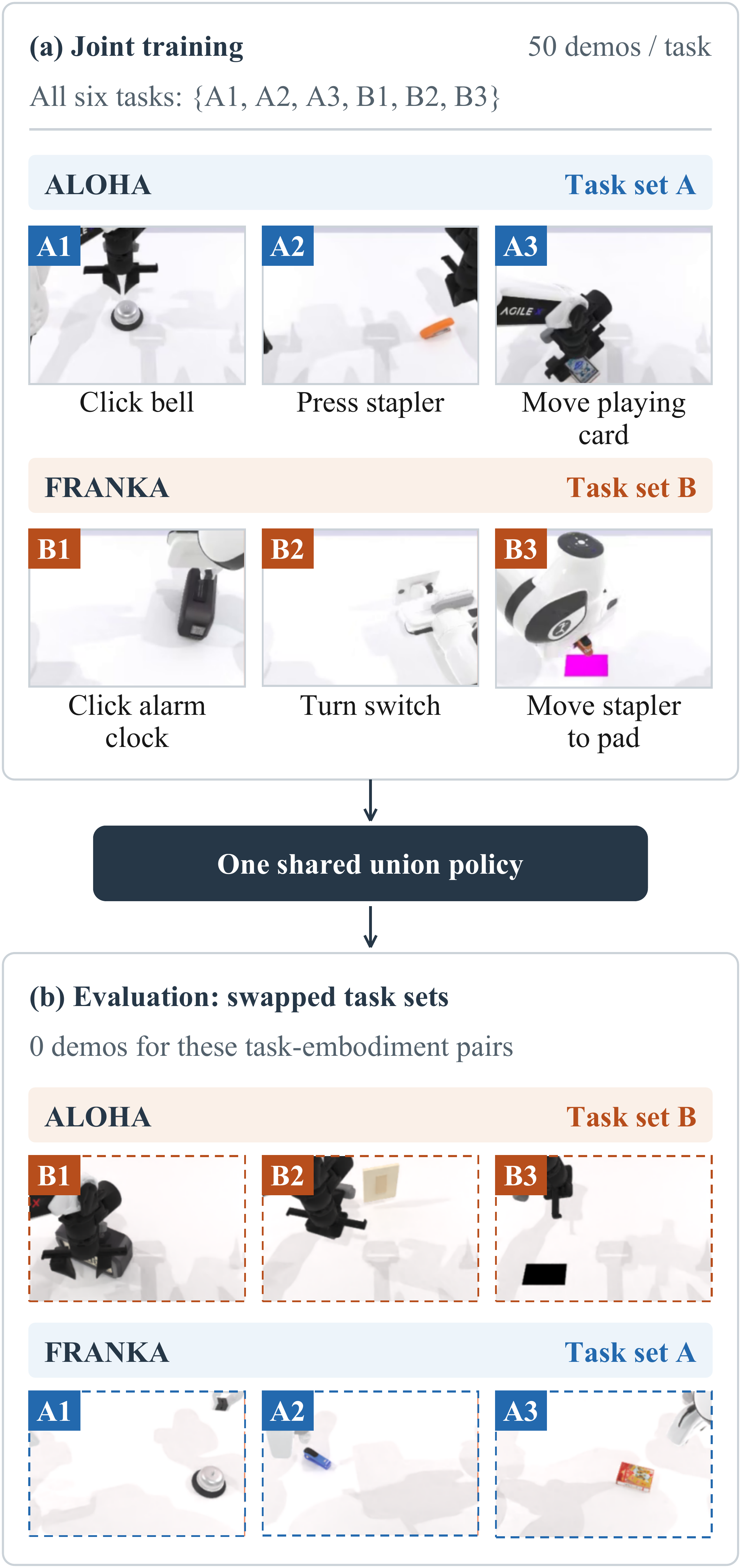}
    \caption{The cross-embodiment task-transfer setting. Each embodiment demonstrates three tasks (top); at evaluation the task sets swap embodiments (bottom): every evaluated embodiment-task pair has zero demonstrations anywhere during training. Bottom thumbnails are frames from actual closed-loop transfer rollouts of our best system.}
    \label{fig:xe_setup}
\end{figure}

Our contributions are:
\begin{itemize}
    \item \textbf{A controlled cross-embodiment experimental setup} on RoboTwin~2.0 \cite{chen2025robotwin2}: using the many embodiments and tasks available in RoboTwin~2.0, we create two disjoint task sets each demonstrated on a unique embodiment, then train a single LAM using the two sets, train a latent action policy, and a decoder per embodiment that map the latent actions to the ground-truth action tasks. By keeping the pipeline parameters fixed except for the LAM, we can isolate the effects of the LAM changes, see Fig.~\ref{fig:xe_setup}.
    \item \textbf{A simple recipe for action-similarity supervision} similar to the global similarity structure loss term of VILA \cite{jeong2026vila} applied to the cross-embodiment setup.
    \item \textbf{A controlled comparison of action target spaces and supervision forms performance} under a fixed policy architecture and hyperparameters, dataset, and training and evaluation protocols we compare raw ground-truth actions (shared padded joint space, and end-effector deltas) vs.\ eight different LAM variants, and an auxiliary action prediction supervision loss \cite{nikulin2025latent} vs.\ an action-similarity supervision loss, on two LAM baselines.
\end{itemize}

\section{Related Works}
In this section, we introduce the related works our study relies on as well as other cross-embodiment methods.

\subsection{Vision-Language-Action Models}
VLA models have appeared in recent years as a way to exploit VLM priors learned on web-scale datasets for robotics tasks \cite{black2024pi0visionlanguageactionflowmodel, intelligence2025pi05visionlanguageactionmodelopenworld, shukor2025smolvlavisionlanguageactionmodelaffordable, kim24openvla, nvidia2025gr00tn1openfoundation}. Recent VLAs are typically VLMs that have been extended with a diffusion or flow-matching decoding head \cite{nvidia2025gr00tn1openfoundation} or expert \cite{black2024pi0visionlanguageactionflowmodel, shukor2025smolvlavisionlanguageactionmodelaffordable}, and fine-tuned on robotic datasets containing action labels using imitation learning. Robot datasets are fragmented between diverse embodiments, leading most policies to adopt a shared padded action space between robots \cite{openx2024}, without any structured latent action. Our ground-truth-action baseline follows exactly this approach, and we show that it is dominated by latent action targets in the low-data cross-embodiment regime.

\subsection{Latent action models from observation}
LAMs recover a compact action interface directly from unlabeled video by jointly training an IDM and an FDM through an information bottleneck on latent action $\bm{z}$ through a low dimension count \cite{nikulin2025latent} or through quantization \cite{schmidt2023lapo}. LAPO \cite{schmidt2023lapo} introduced this approach for reinforcement learning (RL) control benchmarks, Genie \cite{bruce2024geniegenerativeinteractiveenvironments} scaled it into an interactive world model with a small discrete action set, and LAPA \cite{ye2024latentactionpretrainingvideos} took it to robotics by pretraining VLAs on latent actions. AdaWorld \cite{gao2025adaworld} learns a continuous latent action VAE for adaptable world models. A subsequent line of work documents the central weaknesses of purely self-supervised latents: they are sensitive to visual distractors \cite{nikulin2025latent}.

\subsection{Shaping the latent action space with auxiliary losses}
A recent group of works enhances the LAM objective with contrastive or alignment terms that act directly on the latent action space. For instance, LAOM \cite{nikulin2025latent} shows that a small amount of ground-truth action supervision (an auxiliary head decoding the true action from the latent action, discarded after training) substantially improves downstream performance, and CLAM \cite{liang2025clam} grounds a continuous latent with jointly-trained action decoding. VILA \cite{jeong2026vila} adds a weighted InfoNCE plus a structural-alignment loss so that latent actions inferred from different camera viewpoints are pulled together when their future ground-truth action sequences are similar. ConLA \cite{dai2026conla} uses a supervised contrastive loss \cite{khosla2020supcon} grouping latent actions that share a discrete action-class label. MVP-LAM \cite{lee2026mvplam} reaches viewpoint robustness through cross-view reconstruction. These methods leave the latent fully unsupervised or anchor it with \emph{embodiment-specific} ground-truth actions. In our study, we steer the latent only through pairwise \emph{similarities} of action sequences, preserving an abstraction that remains comparable across embodiments similar to the global structure similarity term of VILA \cite{jeong2026vila}. Unlike VILA, our supervision signal targets cross-embodiment transfer (not viewpoint robustness).

\subsection{Cross-embodiment latent action bridges}
Several VLA systems unify embodiments through a shared latent action. IGOR \cite{chen2024igor} compresses visual change into a latent shared across humans and robots. villa-X \cite{chen2025villa0x0} aligns the latent to robot dynamics via proprioceptive forward-dynamics supervision. UniVLA \cite{bu2025univla} extracts task-centric latents in a self-supervised feature space. Moto \cite{chen2024moto} learns hardware-agnostic motion tokens. GO-1 \cite{bu2025agibot_iros} and GR00T~N1 \cite{nvidia2025gr00tn1openfoundation} adopt latent or IDM pseudo-actions inside generalist policies. These shared latent actions are obtained through reconstruction, world modeling, proprioception, or language.

\section{Methodology}
In this section, we introduce the multi-stage pipeline we use to train the LAM, latent policy, and decoder (as represented in Fig. \ref{fig:main_illustration}).

\subsection{The LAM training pipeline}
Let $H>0$ be a time horizon (counted as steps), $\mathcal{D} = \{(\bm{o}_t, \bm{o}_{t+H}, g_t)\}$ a dataset of transitions with an associated goal $g_t$ (e.g.~a text prompt or a goal image), $\mathcal{D_A} = \{(\bm{o}_t, \bm{o}_{t+H}, g_t, \bm{a}_{t:t+H}, e_t)\}$ the action-annotated (with actions $\bm{a}_{t:t+H}$) subset of $\mathcal{D}$ with an embodiment marker $e_t$ specifying which embodiment the actions corresponds to, $f_\text{IDM}(\phi(\bm{o}_t), \phi(\bm{o}_{t+H}))$ an IDM encoding the change $\bm{z}_t$ between the observations called a latent action, $f_\text{FDM}(\phi(\bm{o}_t), \bm{z}_t)$ an FDM outputting a vector $\hat{y}_{t+H}$. In the case where the IDM/FDM pair is trained directly using observation in the pixel-space,  $\phi$ is the identity function, whereas in the case where the IDM/FDM pair, $\phi(\cdot)$ is the visual feature encoder and the FDM outputs feature vector in the visual feature space. For notation simplicity, we will omit the visual encoder $\phi(\cdot)$ in the next sections.

The IDM and the FDM are trained jointly. The IDM encodes the latent action between two observations, while the FDM reconstructs the second observation from the first observation and the latent action:
\begin{equation}
    \bm{z}_t = f_{IDM}(\bm{o}_t, \bm{o}_{t+H}), \qquad
    \hat{\bm{o}}_{t+H} = f_{FDM}(\bm{o}, \bm{z}_t),
\end{equation}
\begin{equation}
    \mathcal{L}_\text{recon} = \mathbb{E}\left[\left\|\bm{o}_{t+H} -  \hat{\bm{o}}_{t+H}\right\|^2_2\right],
\end{equation}

We implement two main versions of the IDM-FDM pair backbones:
\begin{itemize}
    \item \textbf{Variational Auto Encoder (VAE) backbone:} We follow the approach of the LAM component of AdaWorld \cite{gao2025adaworld}, where $f_{\text{IDM}}$ is a spatiotemporal-Transformer encoder over the observations pair that produces a low-dimensional continuous latent action (where $d{=}64$) under a Kullback-Leibler (KL) bottleneck) and ($f_{\text{FDM}}$) is a spatial-only Transformer decoder taht reconstructs the next observation in pixel space (the time horizon is $H{=}1$).
    \item \textbf{Multi-step backbone}: We follow the method described by LAOM \cite{nikulin2025latent}, a convolutional encoder and a wide continuous latent action ($d{=}1024$). The IDM receives  a pair of observations $(\bm{o}_t, \bm{o}_{t+H})$ where the horizon is uniformly sampled $H \sim \mathcal{U}\{1..10\}$, and the FDM predicts the \emph{embedding} of the future observation $\bm{o}_{t+1}$ produced by an exponential moving average copy of the encoder.
\end{itemize}

Once the IDM and FDM are learned, we label the full dataset $\mathcal{D}$ with the frozen IDM, $\bm{z}_t = f_{IDM}(\bm{o}_t, \bm{o}_{t+H})$, and train a latent policy $\pi$ that, given an observation and a task information $\bm{g_t}$ (e.g., a prompt), predicts a \emph{sequence} of $H_z$ consecutive latent actions. We use the flow-matching VLA $\pi_0$ \cite{black2024pi0visionlanguageactionflowmodel}, whose action expert is trained to generate $\hat{\bm{z}}_{t:t+H_z} = \pi(\bm{o}_t, \bm{g}_t)$ with the standard conditional flow-matching objective. Finally, a small per-embodiment decoder $\delta_e$ maps latent actions to the embodiment's actions, trained on its own labeled subset:
\begin{equation}
    \mathcal{L}_{\delta} = \mathbb{E}_{\mathcal{D}_a^{(e)}}\left[\left\|\bm{a}_t -  \delta_e(\bm{z}_t)\right\|^2_2\right].
\end{equation}
The IDM and FDM are discarded after training and at deployment we run $\delta_e(\pi(\bm{o}_t, \bm{g}))$ closed-loop.

\subsection{Action-similarity supervision}
\label{sec:action_similarity_supervision}
Similar to VILA's global structure similarity loss term~\cite{jeong2026vila}, we use the pairwise similarities of the ground-truth action sequences as a supervision signal for the latent actions. The learning mechanism is that of relational knowledge distillation \cite{park2019rkd}, where instead of matching individual targets, the student (the latent action space) matches the pairwise relations of the teacher (the ground-truth action space). On every labeled batch $\{(\bm{o}_t, \bm{o}_{t+H}, \bm{a}_{t:t+H}, e_t)\}_{i=1}^{B}$ we form two $B \times B$ (where $B$ is the batch size) similarity matrices and penalize their difference (see equation~\ref{eq:action_sim_loss}). We use the cosine similarity for both the latent actions and the ground-truth actions.

\textbf{Latent similarities.} We compute the pairwise latent action similarity matrix by $\ell_2$-normalizing the latent vectors then computing the cosine similarity of each pair:
\begin{equation}
    S^z_{ij} = \text{sim}(\bm{z}_i, \bm{z}_j)
\end{equation}
Since the similarities are computed with the latent actions coming out of the IDM, gradients flow back into the IDM during backpropagation.

\textbf{Ground-truth action similarities.} Since $\bm{z}$ encodes a visual \emph{change}, similarity must reflect shared \emph{motion}, not shared absolute pose. Absolute action sequences (e.g., joint-position targets) are therefore converted to per-step deltas $\Delta\bm{a}_t = \bm{a}_{t+1} - \bm{a}_t$ (action spaces that already encode per-step motion, such as end-effector deltas, are used as-is). Deltas are standardized per action dimension over the batch and sequences, flattened over the chunk, $\ell_2$-normalized, and compared with a cosine similarity maximized over small temporal shifts ($\pm2$). With temporal shift $s$,
\begin{equation}
    S^a_{ij} = \max_{d \in \{-s..s\}} \text{sim}\!\left(\bm{A}_i^{(0)},\, \bm{A}_j^{(d)}\right),
\end{equation}
where $\bm{A}_j^{(d)}$ is the sequence of $j$ shifted by $d$ steps; the matrix is symmetrized with an elementwise max. This scores each pair by its best temporal alignment, so equal motions offset by a frame or two still count as similar. This fixed shift window only compensates constant offsets. Alignment methods such as DTW or soft-DTW \cite{cuturi2017softdtw} additionally handle motions executed at different speeds, and we leave them to future work.

\textbf{Loss and masking.} With a pair mask $\mathcal{M} \in \bm{1}^{B\times B}$ whose diagonal is always 0,
\begin{equation}
    \label{eq:action_sim_loss}
    \mathcal{L}_{\text{sim}} = \left\lVert \mathcal{M} \times (S^z - S^a) \right\rVert_{F}^2 ,
\end{equation}
where $||\cdot||_{F}^2$ is the Frobenius norm, and $\times$ the Hadamard product. When action similarities are computed on padded joint-space actions, they are only well-defined \emph{within} an embodiment, and $\mathcal{M}$ can be used to exclude cross-embodiment pairs ($e_i \ne e_j$). With a embodiment-agnostic action space (e.g.~end-effector deltas), the mask can be dropped and the loss term aligns latent actions \emph{across} embodiments directly. The total stage 1 objective is:
\begin{equation}
    \mathcal{L} = \mathcal{L}_{\text{recon}} + \lambda\,\mathcal{L}_{\text{sim}}.
\end{equation}
$\lambda$ is \textit{small} (we test with $\lambda=0.05$); notably its effective strength is backbone-dependent because the reconstruction scale differs across backbones.

\section{Experimental Setup}
\label{sec:setup}
\textbf{Environment and embodiments.} We build on RoboTwin~2.0 \cite{chen2025robotwin2}, a dual-arm SAPIEN-based \cite{xiang2020sapien} manipulation benchmark with interchangeable embodiments observed from an identical head camera. We use two different embodiments: \emph{aloha-agilex} 6-DoF per arm with well behaved inverse kinematics and \emph{franka}, a pair of 7-DoF Franka arms, kinematically redundants. Actions are absolute joint positions encoded in a shared 16 dimensions padded space.

\textbf{Cross-embodiment split.} Each embodiment demonstrates three different tasks with 50 scripted demonstrations per task. The Aloha robot demonstrates the "click bell", "press stapler", and "move playing card away" tasks, while the Franka robot demonstrates the "click alarm clock", "turn switch", and "move stapler pad" tasks.
The LAM is trained on all six tasks' videos from both embodiments, the latent policy is trained on the union of both embodiments' (observation $\to$ latent) pairs; each decoder $\delta$ only on its own embodiment's labels. During evaluation the task sets \emph{swap}: each robot is evaluated closed-loop on the three tasks demonstrated only by the other robot ("cross-embodiment") and on the three tasks demonstrated only by itself ("in-embodiment"). Every cross-embodiment embodiment-task pair has zero demonstrations, so task competence can only have crossed through the shared latent action space and shared latent action policy.

\textbf{Evaluation protocol.} We follow the official RoboTwin~2.0 evaluation protocol: the random seeds for the scene generation during evaluation are different from the data-collection seeds, and each candidate scene is first verified by the scripted expert, and only expert-solvable scenes are evaluated. We report the success rate (SR) over $n{=}50$ episodes per embodiment-task pair, a full benchmark evaluation is 600 episodes.

\section{Experiments}
We organize the study around multiple questions:
\begin{itemize}
    \item {\textbf{Q1: How does action supervision affect transfer?} LAMs are usually unsupervised but, as mentioned earlier, some works have successfully leveraged the few ground-truth action labels via auxiliary losses (prediction head or similarity supervision). We study how these supervision affect the cross-embodiment transfer by comparing unsupervised, prediction head, and similarity supervision.}
    \item {\textbf{Q2: Do the gains persist with fewer labels or a smaller policy?} The benefits of leveraging ground-truth action labels in a LAM are valuable only if these benefits hold when the amount of ground-truth action labeled trajectories is significantly outweighed by the amount of unlabeled trajectories. We study how our recipe react to various amount of labeled data as well as a weaker policy.}
\end{itemize}
After which we propose a simple recipe based on the best performing combination of the study.

\textbf{Compared systems.} Table~\ref{tab:main} compares the use of action labels under the data and evaluation protocol of Sec.~\ref{sec:setup}. Unless stated otherwise, we fix the $\pi_0$ initialization, policy architecture and training hyperparameters. GT-$\pi_0$ predicts padded joint actions directly, and GT-$\pi_0$-EE predicts end-effector deltas. The LAM-based systems use either the AdaWorld VAE backbone or the multi-step LAOM backbone of Sec.~III. Within each backbone, we compare unsupervised pretraining with action-similarity supervision ($+\mathrm{sim}$). On LAOM, we also evaluate an auxiliary action prediction head ($+\mathrm{pred}$)~\cite{nikulin2025latent}. Similarities use joint deltas and within-embodiment pairs by default. EE switches the similarity targets to end-effector deltas; X additionally includes cross-embodiment pairs; MV adds both wrist cameras to the LAM's head-camera input. The EE, X, and MV ablations use the same dimension count for the latent action and the same downstream training protocol. LAOM similarity variants use $\lambda{=}0.05$; the VAE weight is swept separately because objective scales differ across backbones.

\begin{table*}[t]
\centering
\caption{Closed-loop success rate (\%), with $50$ episodes per embodiment-task pair. Own and transfer denote tasks demonstrated by the executing robot and by the other robot; each cell averages three tasks. Aggregate columns average the two embodiments or all twelve pairs. Bold marks column maxima among individual runs; -- denotes unavailable results.}
\label{tab:main}
\small
\setlength{\tabcolsep}{4pt}
\begin{tabular}{lrrrrrrr}
\toprule
 & \multicolumn{2}{c}{Aloha} & \multicolumn{2}{c}{Franka} & \multicolumn{3}{c}{Average} \\
\cmidrule(lr){2-3}\cmidrule(lr){4-5}\cmidrule(lr){6-8}
Configuration & Own & Transfer & Own & Transfer & In-emb. & Transfer & Overall \\
\midrule
GT-$\pi_0$ & 76.7 & 33.3 & 30.7 & 14.7 & 53.7 & 24.0 & 38.9 \\
GT-$\pi_0$-EE & 49.3 & 18.0 & 32.7 & 34.0 & 41.0 & 26.0 & 33.5 \\
\midrule
AdaWorld & 46.7 & 10.7 & 46.0 & 43.3 & 46.4 & 27.0 & 36.7 \\
AdaWorld $+\mathrm{sim}$ ($\lambda{=}0.1$) & 46.0 & 30.7 & 37.3 & 24.0 & 41.7 & 27.4 & 34.5 \\
AdaWorld $+\mathrm{sim}$ ($\lambda{=}0.05$) & 46.0 & 34.0 & 48.0 & 30.7 & 47.0 & 32.4 & 39.7 \\
AdaWorld $+\mathrm{sim}$ ($\lambda{=}0.02$) & 42.0 & 36.7 & 48.0 & 24.0 & 45.0 & 30.4 & 37.7 \\
\midrule
LAOM & 74.0 & 36.0 & 51.3 & 36.0 & 62.7 & 36.0 & 49.3 \\
LAOM $+\mathrm{pred}$ & 70.7 & 46.0 & 60.7 & 22.0 & 65.7 & 34.0 & 49.9 \\
LAOM $+\mathrm{sim}$ & 72.0 & 40.7 & 60.0 & 51.3 & 66.0 & 46.0 & 56.0 \\
LAOM $+\mathrm{sim}$-EE & 75.3 & 55.3 & 65.3 & 50.7 & 70.3 & 53.0 & 61.7 \\
LAOM $+\mathrm{sim}$-EE-X & 77.3 & \textbf{65.3} & 61.3 & 57.3 & 69.3 & 61.3 & 65.3 \\
LAOM $+\mathrm{sim}$-EE-X-MV & \textbf{79.3} & 58.7 & 63.3 & 33.3 & \textbf{71.3} & 46.0 & 58.7 \\
\midrule
ViLA-style (weighted InfoNCE) & 73.3 & 62.7 & \textbf{66.7} & \textbf{64.0} & 70.0 & \textbf{63.3} & \textbf{66.7} \\
ConLA-style (verb-class contrast) & 38.0 & 10.7 & 34.0 & 20.0 & 36.0 & 15.3 & 25.7 \\
\midrule
ViLA-faithful (both loss terms) & 71.3 & 56.0 & 63.3 & 48.7 & 67.3 & 52.3 & 59.8 \\
villa-X (full pipeline) & 72.7 & 19.3 & 37.3 & 25.3 & 55.0 & 22.3 & 38.6 \\
CLAM (co-trained decoder) & 64.0 & 37.3 & 55.3 & 53.3 & 59.6 & 45.3 & 52.5 \\
\bottomrule
\end{tabular}
\end{table*}

\subsection{Q1: How does action supervision affect transfer?}
\label{sec:supervision_results}
\textbf{Latent action targets provide a useful starting point.} Unsupervised LAM pretraining yields $27.0\%$ transfer with AdaWorld and $36.0\%$ with LAOM, compared with $24.0\%$ for GT-$\pi_0$. Directly predicting end-effector deltas gives $26.0\%$ transfer and lowers in-embodiment success rate (SR) from $53.7\%$ to $41.0\%$. Thus, in this setup, changing the raw action coordinates alone does not reproduce the benefit of the multi-step latent action representation.

\textbf{How labels enter pretraining matters.} On LAOM, an action prediction head raises in-embodiment SR from $62.7\%$ to $65.7\%$, but transfer decreases from $36.0\%$ to $34.0\%$. Similarity supervision using the same labels instead reaches $66.0\%$ in-embodiment and $46.0\%$ transfer: a $12.0$-point transfer advantage over action prediction supervision, at similar in-embodiment SR. On the VAE backbone, similarity supervision improves transfer from $27.0\%$ to at most $32.4\%$ across the tested $\lambda$ weights. Its benefit therefore depends on the backbone; the direct comparison with action prediction supervision is established on LAOM.

\textbf{Action labels coordinates and pair selection refine the supervision.} Keeping the embodiment mask and changing only the similarity targets from joint deltas to end-effector deltas raises LAOM transfer from $46.0\%$ to $53.0\%$. Unlike joint-space, end-effector space can remain similar across different arm configurations, providing a plausible explanation for this improvement. Including cross-embodiment pairs then raises transfer to $61.3\%$, with $69.3\%$ in-embodiment and $65.3\%$ overall success. This result is the strongest similarity-matching configuration in Table~\ref{tab:main}: transfer is $2.55\times$ the padded-action baseline and retains $88.5\%$ of the in-embodiment performance. End-effector space is therefore more useful here as relational supervision for latent actions than as a direct policy target.

\textbf{Additional views do not necessarily improve transfer.} Adding wrist views to the LAM increases in-embodiment SR from $69.3\%$ to $71.3\%$, while transfer falls from $61.3\%$ to $46.0\%$. This observation is consistent with reliance on embodiment-dependent visual features, but does not by itself identify the cause. We retain the head-camera-only configuration for the remaining experiments.

\textbf{Alternative losses probe the supervision signal.} With the same end-effector labels and cross-embodiment pairs, a ViLA-style weighted InfoNCE loss~\cite{jeong2026vila} reaches $63.3\%$ transfer and $66.7\%$ overall success, slightly above similarity matching ($61.3/65.3\%$). A ConLA-style verb-class contrastive loss with a time-reversal negative~\cite{dai2026conla,khosla2020supcon} reaches $15.3\%$ transfer and $25.7\%$ overall success (Table~\ref{tab:main}). These are LAM loss adaptations: the ViLA-style variant omits its structural-alignment term and only keep the weighted InfoNCE term, and the ConLA-style variant uses our continuous backbone in place of its VQ tokenizer. They suggest that continuous action-similarity information is useful across loss forms. Q2 evaluates label efficiency for both the weighted InfoNCE and the similarity-matching losses, and policy robustness for the similarity-matching configuration only.

\textbf{Additional baselines.} Table~\ref{tab:main} also reports ViLA-faithful~\cite{jeong2026vila}, which restores structural alignment alongside weighted InfoNCE ($\lambda{=}0.05$ each), reaching $52.3\%$ transfer versus $63.3\%$ for InfoNCE alone in these runs. The villa-X full pipeline~\cite{chen2025villa0x0}, with embodiment-specific context, proprioceptive prediction, a VQ-32 codebook, and a two-expert ACT-style policy without a standalone decoder, reaches $22.3\%$. CLAM~\cite{liang2025clam}, co-training a shared action decoder within LAOM pretraining ($\beta{=}1.0$, no relational loss), reaches $45.3\%$.

\textbf{Decoder fit.}
\label{sec:diagnostics}
The reported stage-four decoder MSE is $0.038$ for AdaWorld, $0.017$ for unsupervised LAOM, $0.015$ with action prediction supervision, and $0.003$ with similarity supervision, a $5\times$ reduction relative to action prediction supervision. This supports improved latent action decodability into ground-truth actions.

\subsection{Q2: Do the gains persist with fewer labels or a smaller policy?}
\label{sec:robustness_results}
\begin{figure}[t]
    \centering
    \includegraphics[width=\linewidth]{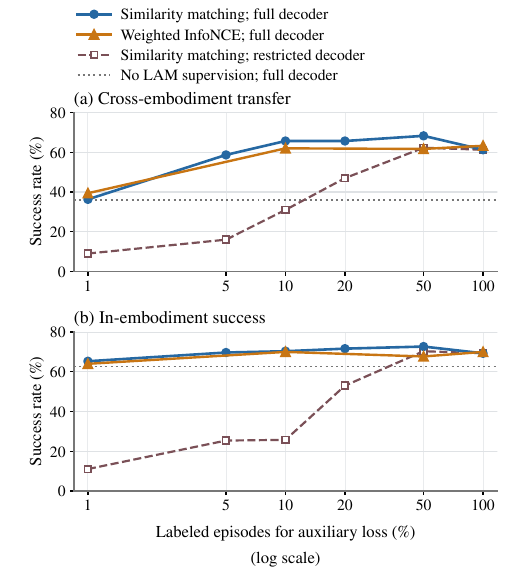}
    \caption{Label efficiency of similarity matching and weighted InfoNCE. A full decoder uses all $150$ labeled episodes per embodiment; a restricted decoder uses the same subset as the auxiliary loss. With full decoders, both objectives retain most of their full-label transfer at $10\%$ labels. The no-LAM-supervision baseline uses a full decoder; the similarity curves share one $100\%$ endpoint. Lines connect evaluated fractions; no intermediate InfoNCE runs are implied.}
    \label{fig:labeleff}
\end{figure}

\textbf{$10\times$ fewer labels for the auxiliary loss.} Table~\ref{tab:efficiency} and Fig.~\ref{fig:labeleff} summarize the labeled-fraction experiments. We vary the fraction reaching the LAM's similarity loss from $1\%$ to $100\%$ ($2$--$150$ labeled episodes per embodiment), keeping the observation data unchanged. We compare training the decoder on the same restricted subset against training it on all $150$ episodes per embodiment. With the full-data decoder, similarity matching reaches $65.7\%$ transfer using only $10\%$ of the auxiliary-loss labels ($15$ episodes per embodiment), compared with $61.3\%$ using all labels. It already reaches $58.7\%$ at $5\%$, whereas $1\%$ gives $36.3\%$, close to the $36.0\%$ baseline with no LAM supervision and a full-data decoder. Thus, the pretraining-supervision benefit is largely retained with $10\times$ fewer labeled episodes.

\begin{table*}[t]
    \centering
    \caption{Aggregate closed-loop success rates (\%) in the label-efficiency study. Each entry is in-embodiment / transfer / 4-cell average. The label fraction is the fraction of episodes whose action labels reach the similarity-supervision loss term.}
    \label{tab:efficiency}
    \small
    \setlength{\tabcolsep}{5pt}
    \begin{tabular}{ccccc}
    \toprule
    LAM labels & Episodes / emb. & Similarity; full decoder & Similarity; restricted decoder & InfoNCE; full decoder \\
    \midrule
    $0\%$   & 0   & 62.65 / 36.0 / 49.3 & -- & -- \\
    $1\%$   & 2   & 65.3 / 36.3 / 50.8 & 11.0 / 9.0 / 10.0 & 64.0 / 39.4 / 51.7 \\
    $5\%$   & 8   & 69.7 / 58.7 / 64.2 & 25.35 / 16.0 / 20.7 & -- \\
    $10\%$  & 15  & 70.35 / 65.7 / 68.0 & 25.7 / 31.0 / 28.3 & 70.0 / 62.0 / 66.0 \\
    $20\%$  & 30  & 71.65 / 65.7 / 68.7 & 53.0 / 47.0 / 50.0 & -- \\
    $50\%$  & 75  & 72.7 / 68.3 / 70.5 & 70.35 / 62.3 / 66.3 & 67.7 / 61.7 / 64.7 \\
    $100\%$ & 150 & 69.3 / 61.3 / 65.3 & 69.3 / 61.3 / 65.3 & 70.0 / 63.3 / 66.7 \\
    \bottomrule
    \end{tabular}
\end{table*}

\textbf{Decoder data remains a limiting factor.} Restricting both similarity supervision and decoder training to $10\%$ gives $31.0\%$ transfer. Training the decoder on all episodes raises this to $65.7\%$, more than doubling success with the same LAM. At $1\%$, the corresponding rates are $9.0\%$ and $36.3\%$; at $50\%$, the gap narrows to $62.3\%$ versus $68.3\%$. At $100\%$, both protocols share the same all-data decoder and the same $61.3\%$ result. The $10\times$ label reduction therefore applies \emph{only} to the auxiliary supervision.

\textbf{The label-efficiency benefit extends to InfoNCE.} With full-data decoders, we also evaluate the ViLA-style weighted InfoNCE loss at $1\%$, $10\%$, $50\%$, and $100\%$ labels. At $10\%$, it reaches $62.0\%$ transfer versus $63.3\%$ at full labels, retaining approximately $98\%$ of its full-label success. Both continuous objectives therefore retain most of their transfer benefit at one-tenth of the auxiliary-loss labels. Their curves differ: at $50\%$, similarity matching reaches $68.3\%$ and InfoNCE $61.7\%$. The non-monotonic results do not establish that fewer labels improve performance or that the objectives are equivalent.

\begin{figure}[t]
    \centering
    \includegraphics[width=\linewidth]{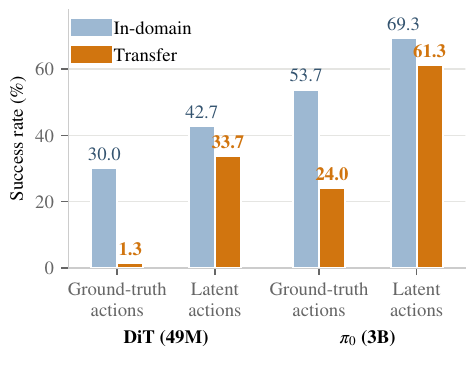}
    \caption{Policy comparison with direct joint-action targets and the selected latent representation. Transfer improves by $37.3$ percentage points for $\pi_0$ and $32.4$ points for the ${\sim}60\times$ smaller Multi-Task DiT~\cite{barreiros2026careful}. The DiT gains over $25\times$ in transfer success and retains $78.9\%$ of its in-embodiment SR. LAM pretraining, annotation, and decoder training are fixed when changing the latent policy.}
    \label{fig:dit2x2}
\end{figure}

\textbf{A smaller policy.} We replace $\pi_0$ with a Multi-Task Diffusion Transformer (DiT)~\cite{barreiros2026careful} with $49$M trainable parameters, approximately $60\times$ smaller than the ${\sim}3$B-parameter $\pi_0$, keeping LAM pretraining, annotation, and decoder training unchanged. For each policy, we compare direct joint-action targets to the reference latent action targets (Fig.~\ref{fig:dit2x2}). The latent actions pipeline increases transfer from $24.0\%$ to $61.3\%$ for $\pi_0$ and from $1.3\%$ to $33.7\%$ for the DiT, gains of $37.3$ and $32.4$ percentage points, respectively. For the DiT, this is more than a $25\times$ increase in transfer success rate; its latent policy retains $78.9\%$ of in-embodiment SR ($33.7/42.7$), compared with $88.5\%$ for $\pi_0$. The selected representation therefore retains its transfer benefit with a much smaller policy, although absolute success is lower.

\subsection{A practical recipe from the study}
\label{sec:recipe}
In this low-data, cross-embodiment study, the results support multi-step LAM pretraining from the shared head-camera view, augmented with action-similarity supervision computed on standardized end-effector-delta sequences as the strongest approach. We use cosine similarity with the temporal sliding window of Sec.~\ref{sec:action_similarity_supervision}, include both within- and cross-embodiment pairs while excluding self-pairs, and set $\lambda{=}0.05$ for the LAOM backbone. The frozen IDM then annotates all demonstrations; a policy is trained to predict the latent actions from all embodiments, and a separately trained set of decoders maps them to each robot's joint actions. This recipe summarizes the controlled comparisons of supervision; the selected latent action pipeline also retains its transfer advantage with smaller policies.

\section{Conclusion}
We presented a study of cross-embodiment transfer in LAMs and identified a practical recipe based on action-similarity supervision that more than doubles cross-embodiment performance compared to the same policy trained on the standard shared padded joint-space directly. We performed extensive ablations, and demonstrated that the action-similarity supervision approach improves cross-embodiment transfer across different LAM baselines, latent policy architecture and parameter count, action representation space, and ground-truth action label availability, and beats the ground-truth action prediction supervision loss approach for cross-embodiment transfer.
Despite our experiments, the embodiments we tested against are all bimanual robot arm with grippers, and testing on embodiments with more heterogeneity remains for future work.
While our approach is promising for the low-data cross-embodiment regime, recent large scale training efforts have shown cross-embodiment emerging as a result of scaling the data \cite{zheng2025x, kareer2025emergence}. We expect that as the amount of data scales, emerging cross-embodiment would close the gap with the selected configuration, but at the cost of large compute requirements.
Because our loss requires only similarities between action sequences, not the actions themselves as targets, it extends naturally to labels such as human wrist trajectories, which we see as a path to training latent action spaces on cheap, heterogeneous supervision.

\section*{Acknowledgement}
We used Claude Code (with Opus 5) to help with the implementation of the method and the integration of the baselines in the experimental framework. Claude Code (with Opus 5) was also used to help make the scripts that generates the plots of this paper. All data was manually inputted to the scripts and no data was generated or outputted by Claude itself. All code was subjected to unit tests and reviewed by the authors.

\addtolength{\textheight}{-2cm}   

\section*{APPENDIX}

\subsection*{Is embodiment identity encoded in the latent actions?}

We examine whether embodiment identity can be predicted from the latent actions produced during stage two. For each evaluated LAM, we sample $10{,}000$ latent actions, with $5{,}000$ from each embodiment, and train an MLP classifier with one hidden layer of $256$ units. We use balanced classes and an $80/20$ training--test split, giving a chance accuracy of $50\%$. As references, we apply the same classification procedure to the ground-truth action sequences in the shared padded joint space and the end-effector-delta space.

The classifier reaches $100\%$ accuracy on every evaluated latent action space. However, the disjoint task split confounds task and embodiment identity: each task is demonstrated by only one embodiment, so a representation that distinguishes tasks can also distinguish embodiments. The ground-truth action references illustrate this limitation. Padded joint actions give $100\%$ accuracy because the two embodiments occupy different coordinates of the shared action space, while end-effector deltas reach $77.9\%$ despite using a shared motion representation. Task-dependent differences in motion and execution speed can therefore provide information about embodiment identity without requiring embodiment-specific action coordinates.

These results show that embodiment identity is recoverable from the evaluated latent actions, but do not isolate whether the classifier relies on embodiment-specific properties or task-dependent motion. Moreover, all evaluated latent action spaces reach the same classification accuracy despite their different cross-embodiment success rates in Table~\ref{tab:main}. Embodiment classification accuracy alone therefore does not explain the transfer differences observed in this setup.

\subsection*{Does the latent action space extend to an unseen embodiment?}

We next examine whether a frozen IDM can annotate demonstrations from an embodiment absent from LAM pretraining. We use a third bimanual embodiment consisting of two 6-DoF UR5 arms with WSG grippers, with $50$ demonstrations per task for all six tasks. We use LAOM+sim-EE-X, the selected similarity-matching configuration in Table~\ref{tab:main}, and train only a new embodiment-specific decoder, $\delta_{\mathrm{UR5}}$. The LAM and latent policy $\pi$ remain frozen, and the latent annotations of the Aloha and Franka demonstrations remain unchanged. We evaluate decoder fit, open-loop replay, and closed-loop success separately.

\textbf{Decoder fit.}
The UR5 decoder trained on annotations from the frozen IDM reaches an MSE of $0.008$, compared with $0.003$--$0.004$ for the two training embodiments and the reported $0.015$ for LAOM+pred. The higher error indicates a less accurate fit on the unseen embodiment, while showing that its ground-truth actions remain decodable from the frozen latent representation.

\textbf{Open-loop replay.}
We annotate a recorded demonstration with the frozen IDM, decode the latent actions, and execute the resulting action sequence in the scene generated with the demonstration's original seed. This evaluation does not use the latent policy. Replaying the recorded ground-truth actions succeeds in $3/3$ episodes, providing a check of the simulation and execution procedure. For UR5, decoded action replay succeeds in $2/3$ episodes on ``click bell'' and $1/3$ on ``click alarm clock'', compared with $3/3$ and $2/3$, respectively, for the training embodiments under the same protocol. Open-loop replay fails on ``turn switch'' for all embodiments, consistent with the sensitivity of precise rotational motions to accumulated execution error. On the two tasks where decoded replay succeeds, UR5 is within one successful episode of the training embodiments. These limited evaluations support the executability of actions decoded from the frozen IDM's annotations of the unseen embodiment.

\textbf{Closed-loop evaluation.}
Combining the frozen latent policy with $\delta_{\mathrm{UR5}}$ gives a success rate of $0.0\%$ on both original task sets, with $n=50$ episodes per embodiment-task pair. All $300$ episodes reach the step limit without simulation failures.

The decoder-fit and replay results show that the frozen IDM and a newly trained decoder can recover executable actions from UR5 demonstrations. This ability does not translate into successful closed-loop control with the frozen latent policy. The gap suggests that policy generalization to observations of the unseen embodiment is a remaining limitation, although these experiments do not isolate visual conditioning as the sole cause. Training the policy across more embodiments, or reducing its reliance on the robot's visual appearance, are directions for future work.

\subsection*{Hyperparameters}

\textbf{VAE backbone (AdaWorld).}
We use a latent dimension of $d=64$, a KL weight of $\beta=10^{-6}$, and pixel-space reconstruction, and train for $60$ epochs.

\textbf{Multi-step backbone (LAOM).}
We use a latent dimension of $d=1024$ and an IMPALA CNN encoder with scale $6$, channel configuration $16/32/32$, and two residual blocks, operating on $64\times64$ observations. The IDM horizon is sampled as $H \sim \mathcal{U}\{1,\ldots,10\}$. The FDM predicts the embedding of $o_{t+1}$ produced by an exponential moving average copy of the encoder, with update coefficient $\tau=0.001$. We use random-shift augmentation and train for $100$ epochs with a batch size of $512$ and a learning rate of $10^{-4}$.

\textbf{Action-similarity supervision.}
For the LAOM backbone, we set the weight of $\mathcal{L}_{\mathrm{sim}}$ to $\lambda=0.05$. The VAE backbone uses the weights reported in Table~\ref{tab:main}. Ground-truth action similarities are computed over sequences of $16$ steps, with a temporal shift window of $s=2$ and per-dimension standardization over the batch and sequence steps. The labeled batch size is $B=128$.

\textbf{Annotation (stage two).}
We use three-frame windows and compute each latent action from the first and last observations with the frozen IDM.

\textbf{Latent policy.}
We initialize $\pi_0$ from the released pretrained checkpoint and train it to predict sequences of $H_z=16$ latent actions. The policy uses head-camera observations at $224\times224$ resolution and is trained for $10$ epochs with a batch size of $8$ and a learning rate of $10^{-4}$. Latent action targets are whitened using statistics computed over the union of both embodiments' annotations.

\textbf{Embodiment-specific decoders.}
Each decoder $\delta_e$ is an MLP with three hidden layers of $256$ units each, trained for $2500$ updates with a learning rate of $3\times10^{-4}$.

\textbf{Evaluation.}
We evaluate $n=50$ episodes per embodiment-task pair. Evaluation scene seeds start at $100000(1+\mathrm{seed})$. Following the official RoboTwin 2.0 protocol, each candidate scene is first checked by the scripted expert, and only expert-solvable scenes are evaluated. The repository documentation provides additional details on the embodiments' construction, degrees of freedom, and action-space layouts.

%
%
%

\bibliography{root}
\bibliographystyle{ieeetr}

\end{document}